\documentclass[a4paper,runningheads]{llncs}
\pdfoutput=1

\usepackage[utf8]{inputenc}
\usepackage[T1]{fontenc}
\usepackage[british]{babel}
\usepackage{lmodern}
\usepackage{microtype}
\usepackage{graphicx}

\newcommand{\onto}[1]{\textsf{#1}}

\usepackage{xcolor}
\definecolor{keywords}{rgb}{0.4,0,0.4}
\definecolor{comments}{rgb}{0.4,0.4,0.4}
\definecolor{strings}{rgb}{0.4,0,0}
\usepackage{listings}
\lstdefinelanguage{SPARQL11}[]{SPARQL}{
    morekeywords={VALUES, INSERT},
    deletecomment=[l]{\#},
    morecomment=[l]{\#\ }
}

\PassOptionsToPackage{hyphens}{url}
\usepackage[pdfusetitle,breaklinks=true,colorlinks,linkcolor=blue,urlcolor=blue,citecolor=blue]{hyperref}
\usepackage{cleveref}
\crefformat{footnote}{#2\footnotemark[#1]#3}

\title{Rail Topology Ontology: A Rail Infrastructure Base Ontology}

\author{Stefan~Bischof\orcidID{0000-0001-9521-8907} \and Gottfried~Schenner\orcidID{0000-0003-0096-6780}}
\authorrunning{S. Bischof and G. Schenner}
\institute{Siemens AG Österreich, Vienna, Austria
\email{\{bischof.stefan,gottfried.schenner\}@siemens.com}}

\makeatletter
\hypersetup{pdfauthor={Stefan Bischof and Gottfried Schenner},pdfkeywords={Rail Topology, Rail Infrastructure, Rail Network, Network Reachability, Ontology, Industrial Knowledge Graph}}
\makeatother

\begin{document}
\maketitle
\begin{abstract}
Engineering projects for railway infrastructure typically involve many subsystems which need consistent views of the planned and built infrastructure and its underlying topology.
Consistency is typically ensured by exchanging and verifying data between tools using XML-based data formats and UML-based object-oriented models.
A tighter alignment of these data representations via a common topology model could decrease the development effort of railway infrastructure engineering tools.
A common semantic model is also a prerequisite for the successful adoption of railway knowledge graphs.
Based on the RailTopoModel standard, we developed the Rail Topology Ontology as a model to represent core features of railway infrastructures in a standard-compliant manner.
This paper describes the ontology and its development method, and discusses its suitability for integrating data of railway engineering systems and other sources in a knowledge graph.

With the Rail Topology Ontology, software engineers and knowledge scientists have a standard-based ontology for representing railway topologies to integrate disconnected data sources.
We use the Rail Topology Ontology for our rail knowledge graph and plan to extend it by rail infrastructure ontologies derived from existing data exchange standards, since many such standards use the same base model as the presented  ontology, viz., RailTopoModel.

\end{abstract}

\begin{description}
\item[Resource Type:] Ontology
\item[Resource URI:] \url{https://w3id.org/rail/topo#}
\end{description}

\section{Introduction}\label{s:intro}

Rail infrastructure is the basis for a significant share of person and freight transportation volume and is considered essential to reach climate goals.\footnote{\url{https://uic.org/com/enews/article/norway-railways-essential-to-achieve-climate-goals}}

Throughout the lifecycle of rail infrastructure, different systems must work together to ensure safe and reliable transport: track vacancy detection, signalling, interlocking, route setting, freight logistics management, timetables, scheduling, ticketing and passenger systems for journey planning and live passenger information. Besides these operational systems, there are tools for engineering the infrastructure as well as systems to assess and monitor the condition of all the various parts of the infrastructure: field devices and the operational systems that control them.
To ensure safety and reliability, all these systems need \emph{consistent} models of the rail infrastructure they depend on. 

Consistency of rail infrastructure data is currently achieved by the implementation of many, often proprietary, data exchange interfaces. Railway infrastructure managers and software vendors intend to reduce the effort of data import and export interfaces by standardizing data exchange formats.

However, data exchange is often not a sufficient approach. On the one hand, new use cases profit from an up-to-date integrated view of the data residing in application-specific databases. Such use cases include asset management, predictive maintenance or global consistency checking. On the other hand, software providers can reduce development effort by relying on standard data models both by reusing domain knowledge of software engineers and omitting the implementation of many proprietary interfaces.

Regardless of the concrete data integration approach, a widely accepted common data model is necessary. Such a common data model is especially useful in an industrial knowledge graph, where data is already represented as RDF/OWL. Additionally, data integration is typically easier to handle in a system using semantic web technologies than in systems based on UML data models and relational databases.

\paragraph{Requirements}

Our ontology engineering process is guided by three types of requirements: competency questions, functional requirements, and adherence to data exchange standards and best practices. The following extends our previously published requirements~\cite{Bischof20}.

To support railway infrastructure engineering, we chose the following two competency questions:
\begin{enumerate}
    \item If a train runs from A to B on the railway network, which infrastructure elements (including their orientation) will be traversed?
    \item What are the possible paths between A and B on the railway network?
\end{enumerate}

We have the following functional requirements regarding modelling the railway domain and adherence to existing data exchange standards: 
(i) represent the (logical) topology of a railway network, 
(ii) contain the main infrastructure elements found in every railway network, such as tracks, switches and signals, 
(iii) the (logical) position and orientation of these infrastructure elements in the railway network must be defined and 
(iv) express logical aggregation, for example, to denote to which station a signal or switch belongs.

When designing and publishing the railway ontology, we intend to increase adoption by following ontology engineering best practices. Specifically, the ontology should be vendor-independent, easily reusable and openly available. The concepts of the ontology must be well documented (especially important in engineering as it must be clear which concepts of the real world correspond to concepts of the ontology) and the ontology should be \emph{minimal}, in the sense that it should contain no aspects not related to the topology of railway networks.

While we aim to stay inside the OWL 2 profiles for efficient reasoning (we are specifically interested in OWL 2 RL for terminological reasoning) modelling the domain accurately is more important.
A more expressive ontology serves also as a more accurate basis for secondary uses. For example, tools to automatically create SHACL shapes for validity checking (e.g., Astrea~\cite{astrea}) or to automatically create REST APIs (e.g., OBA~\cite{oba}) do not depend on OWL 2 profiles and can exploit more expressive input ontologies. 

\par\vspace{\baselineskip}\noindent
Since no previously published ontology fulfils all these requirements, we created the Rail Topology Ontology (RTO). In this paper, we present the RTO and describe our approach to develop and publish it.

The rest of this paper is structured as follows:
Section~\ref{sec:related} sets our work in context to existing related work.
Section~\ref{sec:rto} gives an overview of the RTO and describes the ontology development approach.
Section~\ref{sec:eval} assesses the ontology with respect to the given competency questions and requirements.
Section~\ref{sec:conc} concludes our work, summarizes the paper and gives an outlook on future work.

\section{Related Work}\label{sec:related}


\paragraph{Transportation Ontologies}
Several ontologies containing railway-related concepts have been published in the past. Typically, the concepts and level of detail are determined by the envisioned use case of the ontology. For example, OTN, a general ontology of transportation networks, contains some railway concepts at a level of detail sufficient to describe transportation between railway stations~\cite{lorenz2005ontology}. OTN is also an example of a reusable ontology, as it has been included in the smart-city ontology km4city \cite{bellini2014km4city}. 
Daniele and Pires~\cite{daniele2013ontological} describe an ontological approach to logistics.
Katsumi and Fox~\cite{Katsumi2018OntologiesFT} survey ontologies for transportation research. 
Although ontologies for transportation research often also contain railway concepts, they lack the necessary detail to describe the topology of a railway network at the operating level (e.g., switches, tracks, signals). 

\paragraph{Data Integration Ontologies}
Heterogeneous data in railway system typically arise in two ways. Either different subsystems of the railway system have a slightly different view of the overall system, or there exist country- and vendor-specific views of the same subsystem. For historic reasons, railway signalling is very country-specific, especially in the European Union. Therefore, many ontologies for integrating railway data have been developed by EU-funded research projects and during European initiatives like SHIFT2Rail\footnote{\url{https://shift2rail.org/}} and its predecessors.
As one of the first projects, the InteGRail project \cite{shingler2008rcm} developed ontologies to integrate the major railway subsystems and provide a coherent view of the data. The RaCoOn ontology \cite{tutcher2015enabling} was developed to demonstrate ontology-based data integration of different subsystems and was used in the European Capacity4Rail\footnote{\url{http://www.capacity4rail.eu/}} project. The recent ST4RT \cite{carenini2018semantic} project leverages Shift2Rail Interoperability Framework components to improve interoperability. Their prototype implementations include travel and ticketing applications.
These ontologies typically focus on the interoperability and mapping aspect and not on modelling the topology of the railway network. Additionally, some ontologies were no longer available online and maintained.

\paragraph{Railway infrastructure and signalling ontologies }
Some ontologies have been developed especially for the formal verification of railway infrastructure. These ontologies typically model the railway network in sufficient detail and allow reasoning about the topology of the network. Examples of this class of ontologies are the RI* ontology~\cite{lodemann2013semantic} and the RAISO ontology~\cite{bellini2016raiso}. In principle, these ontologies might answer our competency questions. However, these ontologies are not available online and not aligned to relevant railway standards.

\paragraph{Existing standards and data formats}
The RailTopoModel (RTM) is a (UML-based) model of railway infrastructure that has been standardized as IRS30100 (International Railway Standard) by the UIC (Union internationale des chemins de fer/international union of railways)~\cite{IRS30100}. RailML is an XML-based standard way to exchange railway data. The topological core of RailML is based on the RTM \cite{hlubuvcek2017railtopomodel}. EULYNX\footnote{\url{https://www.eulynx.eu/}} standardizes interfaces and elements of signalling systems and is also based on RTM. Similarly, IFC Rail\footnote{\url{https://www.buildingsmart.org/standards/rooms/railway/ifc-rail-project/}} for building information systems is also aligned with RTM. We selected RTM in its most recent version 1.1\footnote{\label{rtm11}\url{https://uic.org/rail-system/railtopomodel}} as the base resource for our ontology.

\section{Rail Topology Ontology}\label{sec:rto}

For developing the Rail Topology Ontology, we followed the NeOn methodology~\cite{Suarez-Figueroa2012}. In an extensive literature search (documented in the previous section) we found no suitable ontology to satisfy our requirements. We implemented scenario 2 of the NeON methodology: ``Re-engineering Non-Ontological Resources''.

The three activities of the first phase of Scenario 2 of the NeON methodology are search, assessment and selection of an appropriate non-ontological resource. The results of the first two activities are described in the previous section. Based on our requirements, we selected RTM 1.1 as the main resource in the third activity.

\subsection{Resource Engineering}

The second phase, the resource engineering process, consists of the three activities reverse engineering, resource transformation and ontology forward engineering. The result of this phase is the RTO ontology.

\paragraph{Reverse Engineering}
We manually analysed the RTM specification and the UML model before approaching the subsequent activities. 

\paragraph{Resource Transformation}
For transforming the RTM UML model to OWL 2 we adapt the approach of Zedlitz and Luttenberger~\cite{zedlitz2014conceptual} to our requirements. 
The following list summarizes our conversion of UML modelling features to OWL 2:
\begin{description}
\item[UML classes] are converted to OWL classes. UML generalizations are mapped to subclass axioms. Sibling classes (classes with the same direct superclass) are defined as disjoint.
\item[UML attributes] are converted to asymmetric, irreflexive OWL data properties with a single class or a union of classes as domain and one data type in the range. 
\item[UML data types] are converted to the corresponding XML Schema data types for primitive UML types, and to OWL custom data types for UML enumerations.
\item[UML associations] are converted to OWL asymmetric, irreflexive object properties with a single class or a union of classes as domain and one class as the range.
\item[UML aggregations] are converted in the same way as associations. Their additional UML defined constraint -- UML aggregations must be acyclic since instances must not aggregate themselves -- is not expressed
\item[UML compositions] are also converted like aggregations, but they are additionally defined as inverse functional.
\item[UML multiplicities] on ends of UML associations, aggregations and compositions come in (only) three shapes in RTM: 
\begin{description}
\item[\texttt{0..*}] is ignored in the conversion.
\item[\texttt{1..*}] is converted to an existential restriction.
\item[\texttt{1}] is converted to existential restriction and a (qualified) maximum cardinality constraint.  
\end{description}
\end{description}

UML Multiplicity Elements, which are annotations on the ends of UML associations, are not considered by Zedlitz and Luttenberger~\cite{zedlitz2014conceptual}. There exists only one instance of such an element in RTM: the range of the UML association \texttt{elementParts} from \onto{OrderedCollection} to \onto{NetElement} is specified as \texttt{\{ordered\}}. Intuitively, an \onto{OrderedCollection} relates to an \emph{ordered} list of \onto{NetElement}s.
The graph data model of RDF provides no canonical representation of ordered lists; ontology engineers have to choose between several modelling patterns. 
Most prominent, since they are defined by the RDF Schema specification~\cite{rdfs11}, are the different RDF containers (\onto{rdf:Bag}, \onto{rdf:Seq} and \onto{rdf:Alt}) and RDF collections. 
For RTO, we combine RDF collections with the standard ``unordered'' mapping of associations to allow simple unordered retrieval while at the same time keeping the information about the order.
Additionally, the Turtle serialization~\cite{Turtle14} provides an intuitive shorthand syntax for RDF collections using parentheses. 

The class, object and data property URIs are solely derived from the class, association and attribute names, respectively. This allows straightforward transformation of data, other models and user knowledge from RTM. However, we have to (semi-automatically) ensure that this policy creates no inconsistencies. In several cases multiple associations (attributes) were merged into a single object (data) property. In every one of these cases the range of the property is a single class. The domain is then defined as a union of classes. 

\paragraph{Mapping to Upper Ontology}

The upcoming part 14 of the ISO 15926 standard\footnote{once standardized, available at \url{https://standards.iso.org/iso/15926/part14}}, currently a working draft \cite{ISO15926}, serves as an upper ontology for the RTO.  Originally, ISO 15926 aimed at integration of lifecycle data of oil and gas plants. Later the scope widened, and the standard was positioned as a generic industrial upper ontology. Part 14 formalizes the ISO standards concepts in an OWL 2 ontology. 

We manually mapped all RTO top-level classes and some subclasses as well as some object properties to respective classes and object properties of the upper ontology.

\paragraph{Ontology forward engineering}

In this last activity, we define the additional object property \onto{reaches} to simplify  reachability queries necessary for answering the competency questions. We give a more detailed explanation of this property in Section~\ref{ssec:reachability}.

\subsection{Overview of the Rail Topology Ontology}\label{ssec:rto}

This section gives an overview of RTM and RTO. For details on the RTM classes we refer the reader to IRS 30100 standard~\cite{IRS30100} and the RTM 1.1 specification\cref{rtm11}.

\begin{figure}[tb]
    \centering
    \includegraphics[scale=0.188]{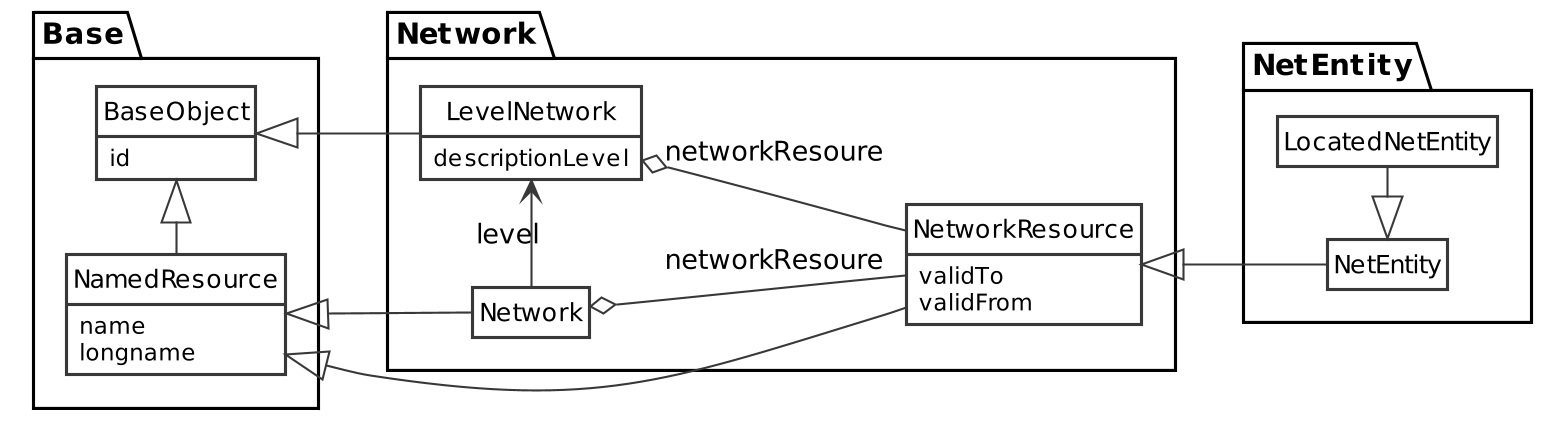}
    \caption{Schematic overview of the Base, Network and NetEntity packages}\label{fig:base-network}
\end{figure}

RTM describes its UML class model in six different packages. We describe the ontology using the same package structure only for didactic reasons, as OWL has no ``package'' construct.

\figurename~\ref{fig:base-network} visualizes the three smallest packages: Base, Network and NetEntity. 

The Base package provides the \onto{BaseObject} to denote objects/instances which are identifiable by some \onto{id}. \onto{NamedResource} is the base class for all instances which can be named. \onto{name} and \onto{longname} are defined as subproperties of \onto{rdfs:label} and \onto{rdfs:comment}, respectively.

The Network package provides concepts to describe rail \onto{Network}s at different levels of detail. One \onto{Network} can have multiple \onto{LevelNetwork}s, each describing the \onto{Network} at the given \onto{descriptionLevel}. The recommended description levels are `macro', `meso' and `micro'. 
Informally, a macro \onto{LevelNetwork} represents nodes (e.g. railway stations) and lines, a meso \onto{LevelNetwork} additionally represents the tracks between nodes, and a micro \onto{LevelNetwork} represents the railway network in detail, i.e., the topology as defined by switches, tracks and crossings.

The NetEntity package includes the base class for \onto{NetEntities}, which are rail infrastructure entities. \onto{LocatedNetEntities} can be located on the rail network using the \onto{Locate} package. The core topology ontology does not contain specific subclasses (e.g. signals, level crossings...) of \onto{Netentities}. These must be provided by additional ontologies, if required.

\begin{figure}[tb]
    \centering
    \includegraphics[scale=0.188]{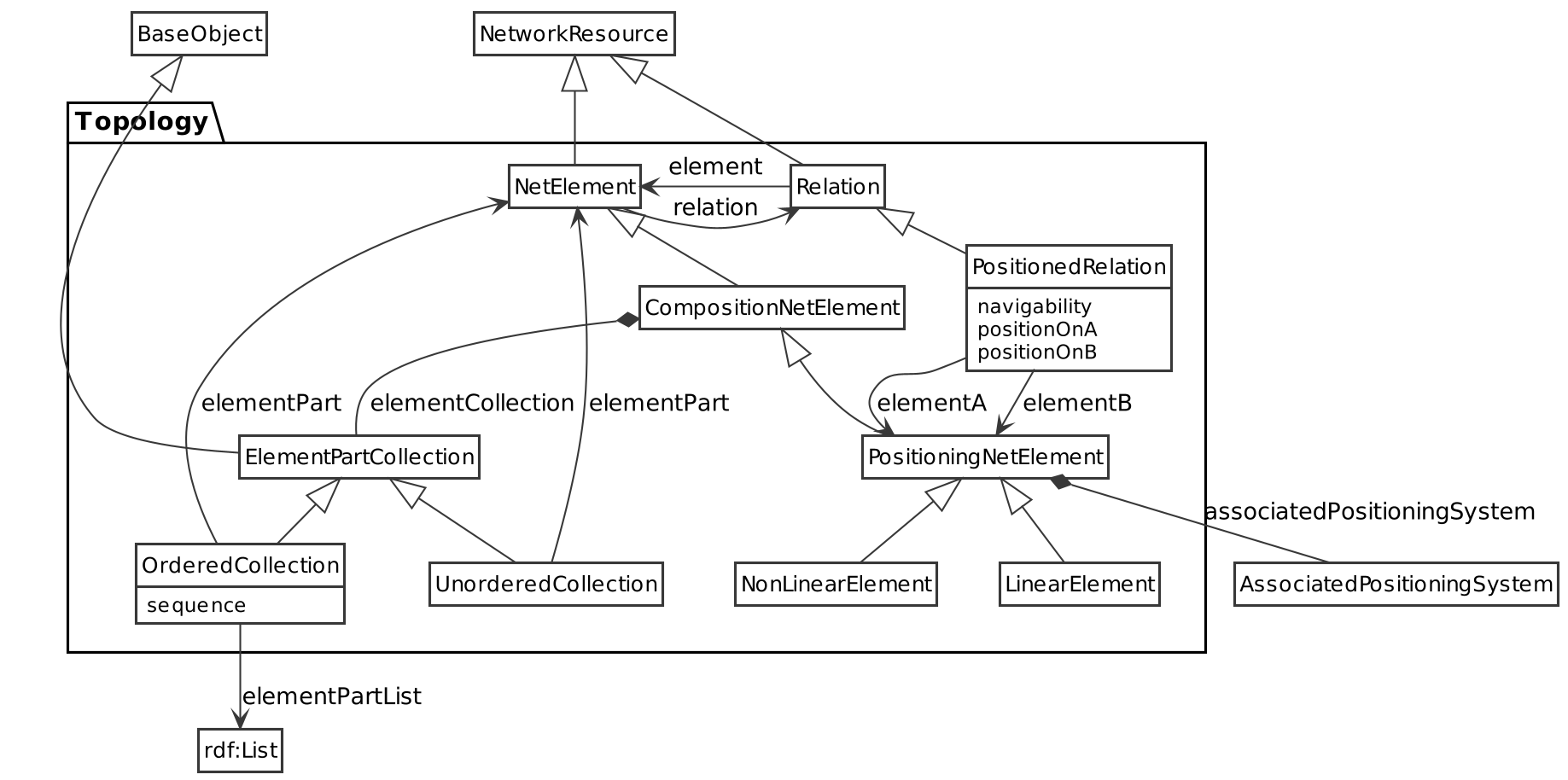}
    \caption{Schematic overview of the Topology package}\label{fig:topology}
\end{figure}

The Topology package, shown in \figurename~\ref{fig:topology}, is used to define the topology of a railway network. The \onto{NetElement} subclasses define segments of the rail network. \onto{PositionedRelation}s define the connection and navigability between \onto{NetElement}s using the properties \onto{navigability}, \onto{positionOnA} and \onto{positionOnB}. 
For example, a \onto{PositionedRelation} with \onto{positionOnA} 0, \onto{positionOnB} 1 and \onto{navigability} \texttt{"AB"} expresses that movement of a train is only possible from A to B between the 0-end of the \onto{elementA} and the 1-end of the \onto{elementB} of the \onto{PositionedRelation}.

The \onto{CompositionNetElement} class aggregates \onto{NetElement}s using the \onto{ElementPartCollection}s. The modelling of the \texttt{\{ordered\}} association end of the \onto{elementPart} association allows users to retrieve the element parts of \onto{OrderedCollection}s and \onto{UnorderedCollection} in a uniform (unordered) manner.

Ordered access is possible via the object property \onto{elementPartList}, which links the \onto{OrderedCollection} to an RDF collection. The following listing shows the Turtle serialization of an \onto{OrderedCollection} \texttt{oc1} with its parts \texttt{ne2}, \texttt{ne1} and \texttt{ne3}, in that order:

\begin{lstlisting}[basicstyle=\footnotesize\ttfamily]
<oc1> a topo:OrderedCollection ;
  topo:sequence 1 ;
  topo:elementPartList ( <ne2>  <ne1>  <ne3> ) ;
  topo:elementPart       <ne1>, <ne2>, <ne3> .
\end{lstlisting}

\begin{figure}[tb]
    \centering
    \includegraphics[scale=0.188]{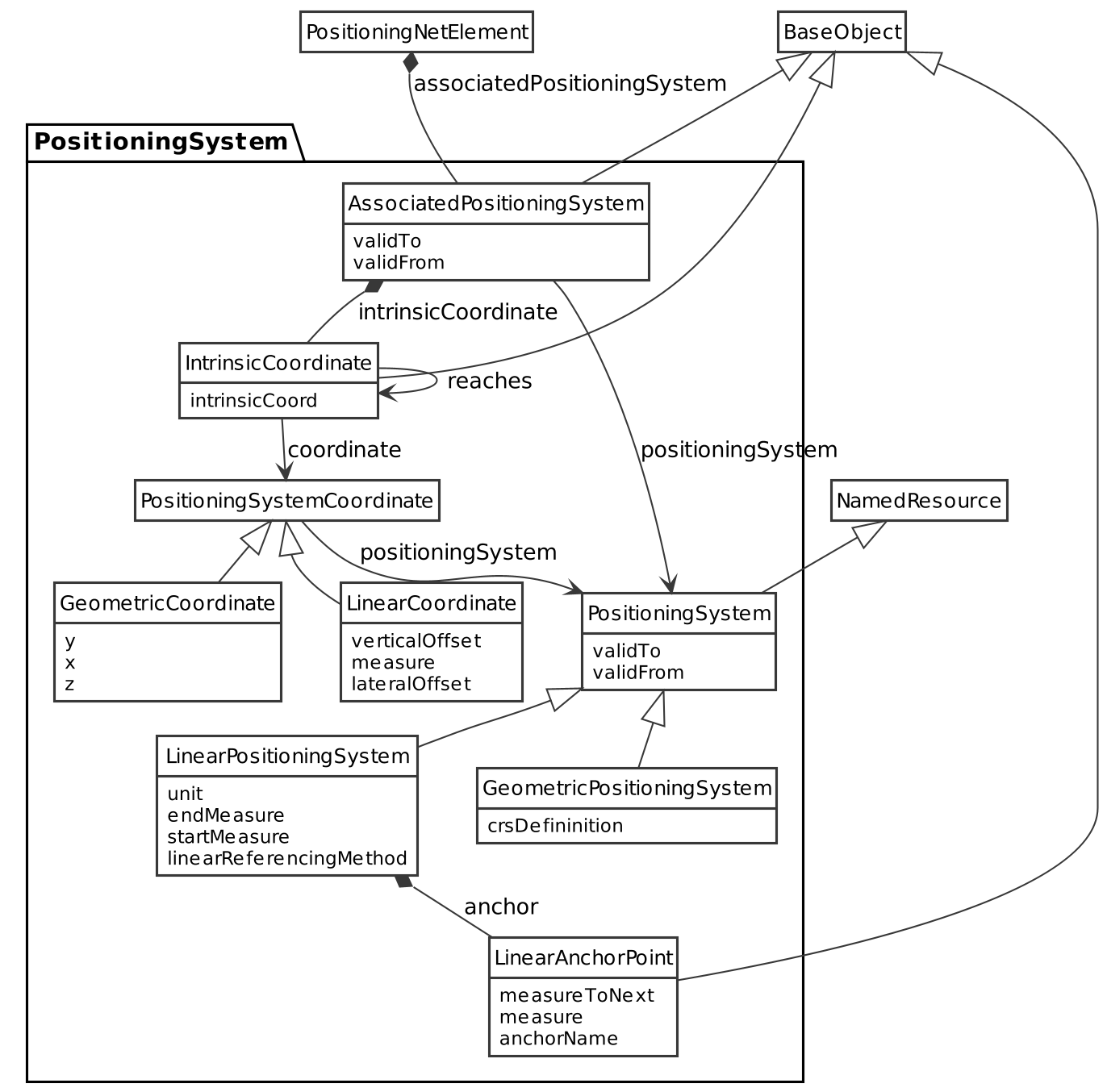}
    \caption{Schematic overview of the PositioningSystem package}\label{fig:positioningsystem}
\end{figure}

\figurename~\ref{fig:positioningsystem} shows the classes for describing positioning systems. Positioning systems are either GPS-based or linear positioning systems that are typically used in line-based railway positioning. \onto{PositioningNetElement}s from the Topology package are assigned positions via the \onto{AssociatedPositioningSystem} class.

\begin{figure}[tb]
    \centering
    \includegraphics[scale=0.188]{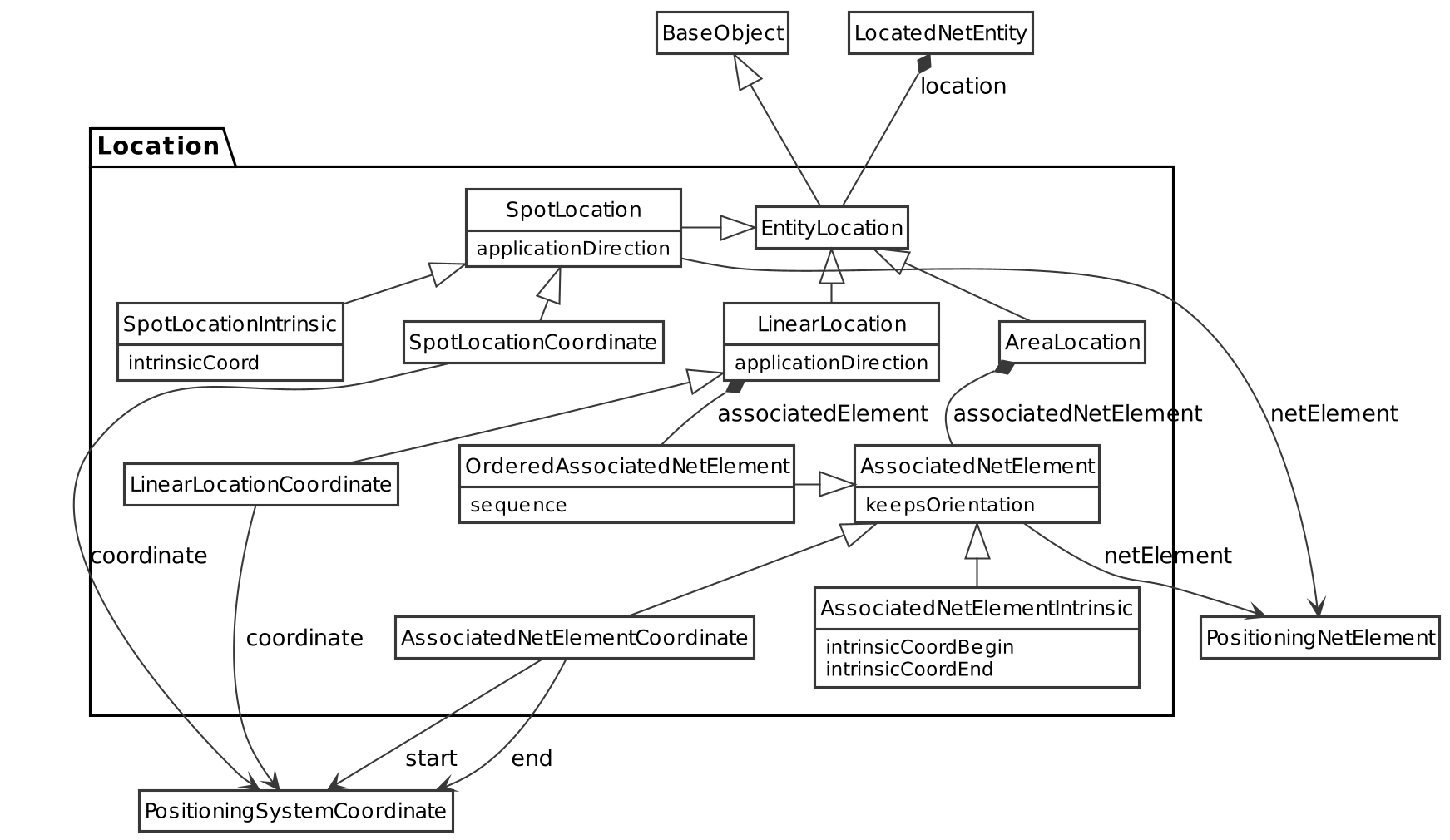}
    \caption{Schematic overview of the Location package}\label{fig:location}
\end{figure}

The Location package (\figurename~\ref{fig:location}) establishes the connection between the \onto{NetEntity} instances and the topology defined by the \onto{NetElement}s. Again, for the details, we refer to the RTO and RTM documentation. The main purpose of the Location package is to assign either a linear location or a spot location to a \onto{NetEntity} by relating it to concepts of the \onto{Topology} and \onto{PositioningSystem} packages. The \onto{applicationDirection} property defines in which direction a \onto{NetEntity} is active, e.g., a railway signal with application direction ``normal'' is only relevant for train movement if the train is moving from 0 to 1 on the corresponding \onto{NetElement} of the signal.

\paragraph{Compatibility with RailTopoModel 1.1}

The ontology was built to be compatible with the UML model of RTM 1.1 to simplify the transition between RTM and RTO. This simplifies not only the transition between RTM and RTO for human users, but also simplifies the development of ontologies based on standards which are themselves based on RTM. 
Contrary to the RTM UML model, all object properties derived from UML associations were defined in singular form instead of the plural used for associations with a maximum cardinality greater than 1 in the RTM UML model. Since object properties link a subject to a single object entity and not collections of entities, their names should be singular.

\subsection{Directed Reachability}\label{ssec:reachability}

Answering reachability queries--i.e., determining which infrastructure elements are reachable by a train moving through the rail network--constitute an important application of the topology ontology. On the micro level of a railway topology, it is often necessary to compute reachability without changing direction. We call this \emph{directed reachability}. Directed reachability can only be defined for \onto{LinearElement}s, because by traversing a \onto{NonLinearElement} we lose the information about the orientation of the train on the element. Unfortunately, writing a SPARQL query to obtain directly reachable \onto{LinearElement}s is non-trivial, since we have to consider the different (local) orientations of the topology elements.

In principle, a train can traverse a \onto{LinearElement} in two directions. In RTM nomenclature, this corresponds to moving from the beginning (\onto{IntrinsicCoordinate} 0) of a \onto{LinearElement} to the end (\onto{IntrinsicCoordinate} 1) or, vice versa. We therefore can define \emph{directed reachability} with a relation \onto{reaches} between \onto{IntrinsicCoordinate}s. Informally, \texttt{:ic1 :reaches :ic2} expresses that a train leaving a \onto{LinearElement} at \onto{IntrinsicCoordinate} \texttt{:ic1} can reach the \onto{IntrinsicCoordinate} \texttt{:ic2} without  changing direction.

\begin{figure}[tb]
  \centering
  \includegraphics[width=.8\textwidth]{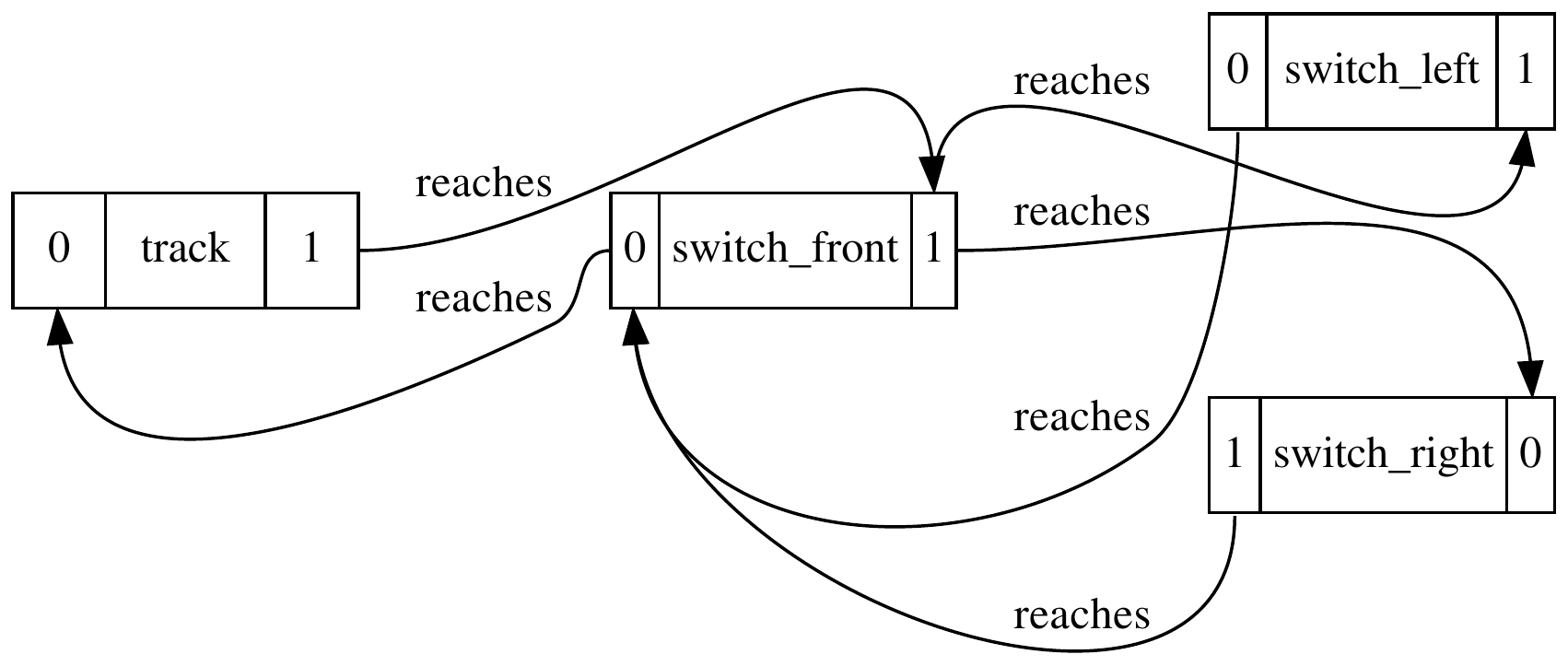}
  \caption{Example: \onto{reaches} relation}\label{fig:reachability-example}
\end{figure}

\figurename~\ref{fig:reachability-example} shows an example consisting of a \onto{NetElement} \texttt{track} representing a track and the \onto{NetElement}s \texttt{switch\_tip}, \texttt{switch\_left} and \texttt{switch\_right} representing the \onto{NetElements} of a railway switch. 
The directed edges correspond to the \onto{reaches} relation, e.g., a train leaving the track at the coordinate \texttt{1} can reach the left and right branches of the switch. A train entering the switch over the left/right branch can only reach the track, but not the other branch of the switch. The \onto{reaches} relation has the following noteworthy properties:
\begin{itemize}
    \item The directions 0/1 are local directions. There is no direct relation to a global direction of the overall rail network or line. Therefore, it is possible for the \onto{reaches} relation to directly go from a \onto{IntrinsicCoordinate} 0 to a \onto{IntrinsicCoordinate} 1, and vice versa (see example).
    \item If the underlying rail network is acyclic, the \onto{reaches} relation consists of two unconnected subgraphs. Each subgraph is a directed acyclic graph and corresponds to one possible train direction.
\end{itemize} 

\begin{lstlisting}[language=SPARQL11,float=tb,caption={Example SPARQL 1.1 query to compute directed reachability for \onto{LinearElement}s},label=l:reach]
PREFIX : <https://w3id.org/rail/topo#>
CONSTRUCT {
  ?icSource :reaches ?icTarget
}
WHERE {
  ?nr a :PositionedRelation ; # PositionedRelation -> NetElements
      ?elemSource ?neSource ; ?posOnSource ?usageSource ;
      ?elemTarget ?neTarget ; ?posOnTarget ?usageTarget .

  # Ensure navigability of PositionedRelation
  { ?nr :navigability "Both" } UNION { ?nr :navigability ?navSingle }

  # Navigate from linear NetElements to intrinsic coordinates
  ?neSource a :LinearElement ; 
    :associatedPositioningSystem/:intrinsicCoordinate ?icSource .
  ?neTarget a :LinearElement ; 
    :associatedPositioningSystem/:intrinsicCoordinate ?icTarget .
    
  # IntrinsicCoordinate -> numeric coordinate
  ?icSource :intrinsicCoord ?iccSource . 
  ?icTarget :intrinsicCoord ?iccTarget . 

  # Leave source NetElement at end 0 or 1
  VALUES (?usageSource ?iccSource ) { 
         (1            1.0 )
         (0            0.0 ) }
    
  # Traverse over target NetElement from end 0 or 1 to the other end
  VALUES (?usageTarget ?iccTarget) { 
         (0            1.0)
         (1            0.0) } 
    
  # navigate NetRelation ?nr either A -> B or B -> A
  VALUES (?elemSource ?elemTarget ?posOnSource ?posOnTarget ?navSingle) { 
         (:elementA   :elementB   :positionOnA :positionOnB "AB")   # A -> B
         (:elementB   :elementA   :positionOnB :positionOnA "BA") } # B -> A
}
\end{lstlisting}

\paragraph{Deriving directed reachability with SPARQL}
In the following, we describe how to derive the \onto{reaches} object property using SPARQL 1.1. SPARQL 1.1 property path expressions can compute undirected reachability of RTO instances. However, directed reachability requires a complicated SPARQL query, taking into account the navigability of the \onto{PositionedRelation}s as well as considering the local direction of the \onto{NetElement}s. The example SPARQL 1.1 \texttt{\color{keywords}CONSTRUCT} query in Listing~\ref{l:reach} computes the \onto{reaches} object property. The \texttt{\color{keywords}VALUES} clauses handle the different orientations of the two \onto{LinearElement}s and the \onto{PositionedRelation}. The materialized \onto{reaches} object property can then be used for more complicated queries like determining the order of elements or the paths between elements. 

\begin{lstlisting}[language=SPARQL11,float=tb,caption={Example query retrieve reachable target \onto{NetElement}s in direction 1.0 of the \texttt{?source} \onto{LinearElement}},label=l:reachquery]
PREFIX topo: <https://w3id.org/rail/topo#>
SELECT ?source ?targetlabel 
WHERE { 
  ?source a topo:LinearElement ; topo:name "switch_front" ; 
      topo:associatedPositioningSystem/topo:intrinsicCoordinate ?ic .
    
  ?ic topo:intrinsicCoord 1.0 ;
      topo:reaches+/^topo:intrinsicCoordinate/
        ^topo:associatedPositioningSystem/topo:name ?targetlabel.
} 
\end{lstlisting}

Listing~\ref{l:reachquery} shows an example SPARQL query to use the \onto{reaches} property in a SPARQL property path expression to retrieve the transitively reachable \onto{LinearElement}s when leaving the \onto{LinearElement} named \texttt{"switch\_front"} at its end 1. For the example given in \figurename~\ref{fig:reachability-example} the SPARQL query would return \texttt{"switch\_left"} and \texttt{"switch\_right"} as bindings for the variable \texttt{?targetlabel}.

Still, some graph properties are hard or impossible (number of paths between two elements) to express in SPARQL. In these cases the \onto{reaches} object property can be converted to a graph representation suitable for a graph library like networkx\footnote{\url{https://networkx.org}}, which provides standard graph algorithms.

\subsection{Ontology Publication}

We publish the RTO under the weak copyleft Mozilla Public License Version 2.0.
The ontology and documentation is available under a permanent URL from W3ID \url{http://w3id.org/rail/topo}. Additionally, the ontology is indexed by Linked Open Vocabularies~\cite{vandenbussche2017linked}.\footnote{\url{https://lov.linkeddata.es/dataset/lov/vocabs/rto}}
The ontology contains metadata annotations for the ontology itself, as well as classes and properties. The ontology documentation was partly generated by WIDOCO~\cite{garijo2017widoco}. The tool also helped publishing the ontology and its documentation following existing best practices: depending on the agent (standard HTTP browser, ontology editing tool) and the used HTTP \texttt{Accept} header, either the documentation or the ontology (in JSON-LD, RDF/XML, N-Triples or Turtle format) is served to the client.

\section{Ontology Assessment and Discussion}\label{sec:eval}

In this section we discuss the ontology's suitability to address the requirements stated in Section~\ref{s:intro}, give arguments on potential adoption and outline a use case scenario.

\paragraph{Requirements}
Both competency questions ask for paths and elements on these paths. RDF and the RTO model can express all the information needed to answer these questions. Although standard SPARQL cannot return paths including the traversed ``edges'' and ``nodes'', custom software can be used to retrieve the necessary information from RTO instance data.

We now discuss whether and to what extent RTO fulfils the functional requirements: 
(i) The topology package of RTO represents the logical topology of  a rail infrastructure network.
(ii) The main infrastructure elements can be represented by using the \onto{LocatedNetEntity} class. Following RTM, we deliberately abstained from explicitly modelling infrastructure elements in RTO. If more fine-grained classes for different infrastructure elements (for example the mentioned tracks, switches and signals) are needed, extensions of that class will be necessary. 
(iii) Modelling position and orientation of infrastructure elements is a core functionality of RTO.
(iv) Aggregation of networks at different levels of abstraction is also a core feature of RTO.

The resulting ontology is vendor-independent since it is based on the international standard RTM 1.1. By following best practices for publishing ontologies, the resource is openly available and should be easy to reuse. 

\paragraph{Expressivity} 
The ontology is contained in OWL 2 DL but not contained in any of the OWL 2 Profiles QL, RL or EL.
The following OWL 2 features used in the UML conversion are not contained in the OWL 2 profiles:
\begin{itemize}
    \item Union of classes in \texttt{rdfs:domain} due to merging of object and data properties with the same name. The alternative of differentiating individual associations and attributes by more precise and verbose naming, for example based on association and domain class name, makes the model more cumbersome to handle: SPARQL queries become harder to read and write due to increased length and, more importantly, users would hardly accept more complicated object and data property names than the ones they know from RTM.
    
    \item Existential restriction resulting from the conversion of the UML cardinalities. There is no alternative way to express a (minimum) cardinality of 1 in the three OWL 2 Profiles. However, cardinalities are necessary and useful in many scenarios (for example for the mentioned tools for automatic SHACL shapes or REST API creation).
    
    \item Definition of custom data types from the mapping of UML enumerations and their use in the range of data properties. 
    
    \item The use of finite data types such as Boolean, float or double, which is forbidden in OWL 2 QL. RTO inherits these data types from RTM. The remedy of falling back to \texttt{xsd:string} is considered infeasible since it would severely limit data validation (for example via SHACL).
\end{itemize}
When selecting a reasoner for RTO data, users must be aware of these limitations.

\paragraph{Use Case}

We use the RTO in our in-house rail knowledge graph. Its main data sources are different rail infrastructure engineering tools. The knowledge graph provides global access to the otherwise disconnected engineering data. 

To set this isolated data into a spatial and network context, we integrate data supplied by the European Union Agency for Railways (ERA) with their EU-wide rail infrastructure database ``Registers of Infrastructure'' (RINF)\footnote{\url{https://www.era.europa.eu/registers_en\#rinf}}. 
Rail infrastructure managers of all EU countries are obligated to provide data about their infrastructure, which is then published by RINF. With RTO, we can access data of these sources using one common model independent of the source data. 

\paragraph{General}

To base the ontology on an existing standard avoids ``reinventing the wheel'' and allows a subject-matter expert for the existing standard to identify the common concepts easily. Also, it is unrealistic to expect that a newly created railway ontology without any relation to existing standards will be adopted by the community. 

On the other hand, as the standard is based on UML the model--especially the class hierarchy and naming conventions--might feel unfamiliar to knowledge engineers used to OWL ontologies.

Although the RTM way of modelling the railway topology might not be intuitive for the non railway expert, it has been well documented why the complexity is necessary \cite{gely2010multi}. A basic knowledge of the railway domain is necessary to use the ontology effectively. Other aspects like the use of the Location package can be overwhelming for the beginner. This is one of the reasons why we will try to accompany the ontology with corresponding SPARQL queries and instance data to illustrate the use of the ontology.

Deriving the ontology from an existing standard enables easier data exchange with existing systems. As the existing standard is based on UML, data exchange is not entirely without effort because of the required mapping between the RTM UML model and the corresponding XML schema (RailML) for serialization. This additional mapping complexity would not exist in standards directly based on RDF/OWL.

For sustainability of the ontology, we maintain the RTO internally through its use in our internal rail knowledge graph. In this knowledge graph, we integrate and provide access to data from several rail infrastructure data sources. The RTO serves as a core schema for this integration. Externally, we further develop the RTO within the new RailML semantic modelling working group. 

\section{Summary and Future Work}\label{sec:conc}

With the RTO we address the first of our previously identified challenges in creating a rail knowledge graph~\cite{eswc2019}: the lack of a standard ontology for rail infrastructure engineering. We believe that standard modular ontologies are a prerequisite for the adoption of semantic technologies in industry. Also, we have experienced in the past that the lack of standard (UML) data models created a lot of inefficiencies and effort for data integration, even when the data models were only slightly incompatible.
This ontology should foster collaboration between the semantic web community and the railway community. 


Specializations of the \onto{LocatedNetEntity} class with different types of rail infrastructure elements are necessary. These include switches, signals and tracks on the micro level and operating points, lines and section of lines on the macro level.
Due to the tight alignment to RTM, the RTO can be a good basis for the development of more specialized rail infrastructure ontologies derived from standard specifications based on RTM, specifically for developing ontologies derived from RailML, EULYNX or IFC Rail.
This approach helps to accelerate ontology development and to integrate data of these different formats. Furthermore, the ontology could serve as a starting point for mapping between railway infrastructure ontologies.
In the new RailML ontology working group, we are working on a RailML-based infrastructure ontology.

When integrating data in a knowledge graph, instance linking (or alignment) is another important task. The RTO topology data have special challenges concerning this task. To accommodate a broad range of use cases, RTM (and thus RTO) leaves a lot of modelling freedom, which makes instance alignment harder. We are investigating different approaches to this problem and for improving the data integration process.

\subsubsection{Acknowledgements}
We thank the anonymous reviewers for their constructive feedback. For authoring the ontology and documentation we used Prot{\'e}g{\'e} 5.5~\cite{musen2015protege}, Widoco~\cite{garijo2017widoco} and the OOPS! ontology scanner~\cite{poveda2014oops}. To prepare the figures in this paper we used PlantUML 
and Graphviz.

\bibliographystyle{splncs04}
\bibliography{references.bib}

\begin{thebibliography}{10}
\providecommand{\url}[1]{\texttt{#1}}
\providecommand{\urlprefix}{URL }
\providecommand{\doi}[1]{https://doi.org/#1}

\bibitem{Turtle14}
Beckett, D., Berners-Lee, T., Prud'hommeaux, E., Carothers, G. (eds.): RDF 1.1
  Turtle. W3C Recommendation (2014), \url{https://www.w3.org/TR/turtle/}

\bibitem{bellini2016raiso}
Bellini, P., Nesi, P., Zaza, I.: {RAISO}: Railway infrastructures and signaling
  ontology for configuration management, verification and validation. In: 10th
  International Conference on Semantic Computing. pp. 350--353. IEEE (2016).
  \doi{10.1109/ICSC.2016.94}

\bibitem{bellini2014km4city}
Bellini, P., Benigni, M., Billero, R., Nesi, P., Rauch, N.: Km4city ontology
  building vs data harvesting and cleaning for smart-city services. Journal of
  Visual Languages \& Computing  \textbf{25}(6),  827--839 (2014).
  \doi{10.1016/j.jvlc.2014.10.023}

\bibitem{eswc2019}
Bischof, S., Schenner, G.: Challenges of constructing a railway knowledge
  graph. In: The Semantic Web: ESWC 2019 Satellite Events. pp. 253--256 (2019).
  \doi{10.1007/978-3-030-32327-1\_44}

\bibitem{Bischof20}
Bischof, S., Schenner, G.: Towards a railway topology ontology to integrate and
  query rail data silos. In: Proceedings of the Demos and Industry Tracks of
  the 19th International Semantic Web Conference (ISWC'20). No.~2721 in \ceur\
  Workshop Proceedings, CEUR-WS.org (2020),
  \url{http://ceur-ws.org/Vol-2721/paper588.pdf}

\bibitem{rdfs11}
Brickley, D., Guha, R. (eds.): RDF Schema 1.1. {W3C} Recommendation (2014),
  available at \url{https://www.w3.org/TR/rdf-schema/}

\bibitem{carenini2018semantic}
Carenini, A., Comerio, M., Celino, I.: Semantic-enhanced national access points
  to multimodal transportation data. In: ISWC 2018 Posters \& Demonstrations,
  Industry and Blue Sky Ideas Tracks. No.~2180 in \ceur\ Workshop Proceedings,
  CEUR-WS.org (2018), \url{http://ceur-ws.org/Vol-2180/paper-09.pdf}

\bibitem{astrea}
Cimmino, A., Fern{\'{a}}ndez{-}Izquierdo, A., Garc{\'{\i}}a{-}Castro, R.:
  Astrea: Automatic generation of {SHACL} shapes from ontologies. In:
  Proceedings of the 17th International Conference {ESWC} (ESWC'20). Lecture
  Notes in Computer Science, vol. 12123, pp. 497--513. Springer (2020).
  \doi{10.1007/978-3-030-49461-2\_29}

\bibitem{daniele2013ontological}
Daniele, L., Pires, L.F.: An ontological approach to logistics. Enterprise
  interoperability, research and applications in the service-oriented
  ecosystem, IWEI  \textbf{13},  199--213 (2013). \doi{10.1002/9781118846995}

\bibitem{garijo2017widoco}
Garijo, D.: Widoco: a wizard for documenting ontologies. In: Proceedings of the
  16th International Semantic Web Conference (ISWC'17) Part {II}. Lecture Notes
  in Computer Science, vol. 10588, pp. 94--102. Springer (2017).
  \doi{10.1007/978-3-319-68204-4\_9}

\bibitem{oba}
Garijo, D., Osorio, M.: {OBA:} an ontology-based framework for creating {REST}
  {API}s for knowledge graphs. In: Proceedings of the 19th International
  Semantic Web Conference (ISWC'20) Part {II}. Lecture Notes in Computer
  Science, vol. 12507, pp. 48--64. Springer (2020).
  \doi{10.1007/978-3-030-62466-8\_4}

\bibitem{gely2010multi}
G{\'e}ly, L., Dessagne, G., Pesneau, P., Vanderbeck, F.: A multi scalable model
  based on a connexity graph representation. Computers in Railways XII
  \textbf{1},  193--204 (2010). \doi{10.2495/CR100191}

\bibitem{hlubuvcek2017railtopomodel}
Hlubu{\v{c}}ek, A.: {RailTopoModel} and {RailML} 3 in overall context. Acta
  Polytechnica CTU Proceedings  \textbf{11},  16--21 (2017).
  \doi{10.14311/APP.2017.11.0016}

\bibitem{Katsumi2018OntologiesFT}
Katsumi, M., Fox, M.: Ontologies for transportation research: A survey.
  Transportation Research Part C-emerging Technologies  \textbf{89},  53--82
  (2018). \doi{10.1002/9781118846995.ch21}

\bibitem{lodemann2013semantic}
Lodemann, M., Luttenberger, N., Schulz, E.: Semantic computing for railway
  infrastructure verification. In: 7th International Conference on Semantic
  Computing. pp. 371--376. IEEE (2013). \doi{10.1109/ICSC.2013.69}

\bibitem{lorenz2005ontology}
Lorenz, B., Ohlbach, H.J., Yang, L.: Ontology of transportation networks.
  Project deliverable, REWERSE project (2005),
  \url{http://rewerse.net/deliverables/m18/a1-d4.pdf}

\bibitem{musen2015protege}
Musen, M.A.: The prot{\'e}g{\'e} project: A look back and a look forward. AI
  matters  \textbf{1}(4),  4--12 (2015). \doi{10.1145/2757001.2757003}

\bibitem{poveda2014oops}
Poveda-Villal{\'o}n, M., G{\'o}mez-P{\'e}rez, A., Su{\'a}rez-Figueroa, M.C.:
  {OOPS! (OntOlogy Pitfall Scanner!): An On-line Tool for Ontology Evaluation}.
  International Journal of Semantic Web and Information Systems (IJSWIS)
  \textbf{10}(2),  7--34 (2014). \doi{10.4018/ijswis.2014040102}

\bibitem{shingler2008rcm}
Shingler, R., Fadin, G., Umiliacchi, P.: From rcm to predictive maintenance:
  The integrail approach. In: 4th International Conference on Railway Condition
  Monitoring. IET (2008). \doi{10.1049/ic:20080324}

\bibitem{Suarez-Figueroa2012}
Su{\'{a}}rez{-}Figueroa, M.C., G{\'{o}}mez{-}P{\'{e}}rez, A.,
  Fern{\'{a}}ndez{-}L{\'{o}}pez, M.: The {NeOn} methodology for ontology
  engineering. In: Ontology Engineering in a Networked World, pp. 9--34.
  Springer (2012). \doi{10.1007/978-3-642-24794-1\_2}

\bibitem{tutcher2015enabling}
Tutcher, J., Easton, J.M., Roberts, C.: Enabling data integration in the rail
  industry using rdf and owl: The {RaCoOn} ontology. ASCE-ASME Journal of Risk
  and Uncertainty in Engineering Systems, Part A: Civil Engineering
  \textbf{3}(2) (2015). \doi{10.1061/AJRUA6.0000859}

\bibitem{IRS30100}
UIC: {RailTopoModel} -- railway infrastructure topological model. Standard,
  International Railway Solution IRS 30100:2016, International Union of Railway
  (UIC) (2016)

\bibitem{vandenbussche2017linked}
Vandenbussche, P.Y., Atemezing, G.A., Poveda-Villal{\'o}n, M., Vatant, B.:
  Linked open vocabularies (lov): a gateway to reusable semantic vocabularies
  on the web. Semantic Web Journal  \textbf{8}(3),  437--452 (2016).
  \doi{10.3233/SW-160213}

\bibitem{ISO15926}
Walther, D., Klüwer, J.W., Martin-Recuerda, F., Waaler, A., Lupp, D., Brandt,
  M.M., Grimm, S., Koleva, A., Kahn, M., Hella, L., Sandsmark, N.: Industrial
  automation systems and integration--integration of life-cycle data for
  process plants including oil and gas production facilities--part 14:
  Industrial top-level ontology. Deliverable, READI project (2020),
  \url{https://readi-jip.org/wp-content/uploads/2020/10/ISO_15926-14_2020-09-READI-Deliverable.pdf},
  working Draft (WD) Proposal for ISO 15926-14:2020(E)

\bibitem{zedlitz2014conceptual}
Zedlitz, J., Luttenberger, N.: Conceptual modelling in {UML} and {OWL}-2.
  International Journal on Advances in Software  \textbf{7}(1 \& 2),  182--196
  (2014), \url{http://www.iariajournals.org/software/}

\end{thebibliography}
\end{document}